# SPINN: An Optimal Self-Supervised Physics-Informed Neural Network Framework

Reza Pirayeshshirazinezhad

Texas A&M University, TX 77553, USA

**Abstract**

A surrogate model is developed to predict the convective heat transfer coefficient of liquid sodium (Na) flow within rectangular miniature heat sinks. Initially, kernel-based machine learning techniques and shallow neural network are applied to a dataset with 87 Nusselt numbers for liquid sodium in rectangular miniature heat sinks. Subsequently, a self-supervised physics-informed neural network and transfer learning approach are used to increase the estimation performance. In the self-supervised physics-informed neural network, an additional layer determines the weight the of physics in the loss function to balance data and physics based on their uncertainty for a better estimation. For transfer learning, a shallow neural network trained on water is adapted for use with Na. Validation results show that the self-supervised physics-informed neural network successfully estimate the heat transfer rates of Na with an error margin of approximately +8%. Using only physics for regression, the error remains between 5% to 10%. Other machine learning methods specify the prediction mostly within +8%. High-fidelity modeling of turbulent forced convection of liquid metals using computational fluid dynamics (CFD) is both time-consuming and computationally expensive. Therefore, machine learning based models offer a powerful alternative tool for the design and optimization of liquid-metal-cooled miniature heat sinks.

## 1. Introduction

The average Nusselt number ($Nu_{ave}$) is a dimensionless parameter that describes the average heat transfer of a fluid flow in a heat exchanger, crucial for understanding and optimizing various industrial processes. Traditional methods for estimating the Nusselt number, such as experimental measurements and high-fidelity CFD simulations, are often expensive and time-consuming. Moreover, experimental data for the heat transfer rate of liquid metal flows in heat exchangers operating under high temperatures and heat loads are scarce. A fast and reliable predictive model for $Nu_{ave}$ of commonly used coolants can be a powerful tool for engineers in the primary design and optimization stage of heat exchangers. This need is especially critical for applications involving liquid metals under extreme working conditions where experimental data are often limited. This paper uses 87 data points from high-fidelity CFD simulations in machine learning (ML) algorithms to estimate the $Nu_{ave}$ of Na in rectangular miniature heat sinks with high accuracy and efficiency.

ML is used in CFD applications to predict the fluid characteristics. In [1], neural network (NN), adaptive neuro-fuzzy inference system (ANFIS), and least squares support vector machine (LSSVM) models are used to estimate the $Nu_{ave}$ of water-based nanofluid flows for a small dataset. Other applications of ML in estimating the fluid characteristics include the estimation of air dew point temperature [2], in the prediction of thermal conductivity [3], and prediction of the relative viscosity of nanofluids [4]. In [5], convolutional neural networks (CNNs) are used with field images as training datasets for steady and unsteady flow field predictions, requiring a large image dataset in the order of 500,000 images. In [6], advancements and challenges in the application of physics-informed machine learning for fluid mechanics are reviewed, focusing on the integration of domain knowledge into ML algorithms to improve data efficiency and prediction stability in complex turbulent flow simulations. In [7], physics-informed time series NN is used to accelerate CFD simulations by inputting two field images to predict the future. Transfer learning showed 1.8 times higher speed in accelerating their CFD simulations [8]. Transfer learning, a technique leveraging knowledge acquired from a source domain for application in a target domain, offers significant benefits for ML processes as detailed by [9]. This method effectively reduces the need for extensive labeled datasets and can enhance the accuracy of machine learning models, decreasing the number of training epochs required and thus increasing computational efficiency. In [10], transfer learning is used for CNNs to address the issue of small dataset. In [11], transfer learning is used to incorporate the experimental data into updating the model predictive optimal controller for particle accelerator.

Per the objectives and contributions detailed in [12], $Nu_{ave}$ of liquid sodium (Na) is modeled using analytical multivariate regression methods with an estimation error of approximately 5%. The error is defined as the maximum absolute difference between predictions and data. In [13], multivariate regression methods are compared with ML methods, with ML providing higher accuracy for prediction. This research develops ML regression algorithms for predicting the $Nu_{ave}$ of liquid sodium (Na). The decision factors for defining $Nu_{ave}$ are the geometrical property of the pipe, Pe (Peclet number), Dh (hydraulic diameter), $\alpha$ (thermal diffusivity). Computational data [12] for Nusselt numbers of liquid sodium in laminar and turbulent flow regimes within stainless steel (SS-316) rectangular miniature heat sinks are used in this work to train ML models. Na-cooled miniature heat sinks have ranges of 0.143-1, 1 mm – 3 mm, and 20-550 $W/cm^2$ for aspect ratios, hydraulic diameters, and heat loads respectively. The Na flow Reynolds number ranges between 500 to 20,000, covering both laminar and turbulent regimes [12]. In this research, ML techniques are introduced to predict $Nu_{ave}$ of liquid sodium (Na) considering the limited number of CFD simulations. Modeling turbulent forced convection of liquid metals with low Prandtl numbers is a computationally expensive task. For the forced convection of liquid sodium in miniature heat sinks, fine mesh elements in the order of 10-100 microns are usually required at the fluid-solid interfaces to keep maximum wall $y^+$ below 1 and accurately estimate temperature and velocity gradients [12,17]. It takes an average of three days to produce one CFD data point for the $Nu_{ave}$ of liquid sodium within miniature heat sinks. The hardware used for the CFD simulation is an Intel Xeon CPU with 20 cores at 3.4 GHz and 64 GB RAM. Due to the high computational cost, the total number of generated CFD data points for the liquid sodium $Nu_{ave}$ is limited to 87 in this work.

Due to the small dataset (87 data points), NN-based methods and Kernel-based learning algorithms, including support vector regression (SVR) and Gaussian process (GP), are introduced for learning and estimating the Nusselt number. Kernel-based learning algorithms and NN show good performance for small datasets [13]. In [14], an optimal structure of shallow NN is developed for estimating water characteristics using genetic algorithm (GA) optimization. To obtain a better extrapolation and accuracy for the NN, the underlying physics of the fluid can be integrated into NN using physics-informed neural network (PINN) approach. The physics equations are evaluated in the loss function using the input to the NN. The loss function in the NN is composed of the prediction error plus a weighted coefficient for the physics to weigh the physics prediction error. In another approach for a generalized better extrapolation, transfer learning is used, where the NN for water is used for training Liquid Sodium (Na)'s NN with transfer learning methods. The given ML algorithm structure is designed by hyperparameter optimization. The optimization algorithm is chosen based on the ML algorithm. Mean Absolute Percentage Error (MAPE) is used to evaluate the performance of the ML algorithm and the minimization of the ML optimization algorithms. To Validate the ML algorithms and the hyperparameter optimization, a 10-fold cross-validation is used to measure the MAPE. A small holdout dataset is used to measure the variation between CFD data and the prediction. A Monte-Carlo simulation is used, where 500 evaluations of the NN-based algorithms measure the variation in prediction, MAPE, and epochs. MAPE, the variation in prediction, the variation in MAPE, and epochs are used as metrics for validation of NN performance. In [15,16], Monte Carlo cross-validation (MCCV) is used in a healthcare context as a resampling approach to evaluate machine learning models, especially when sample size is limited. In [18,19], hyperparameter optimization and Monte Carlo simulation is used for ML algorithms to satisfy and validate the criteria of a distributed telescope in space.

The unique contributions in this work are:
1) Developing ML algorithms for estimating the $Nu_{ave}$ of Liquid Sodium (Na).
2) Development of self-supervised PINN for heat transfer study.
3) Uncertainty quantification of ML and physics-based regression models.
4) Development of transfer learning to transfer the shallow NN of another fluid (water) to Liquid Sodium (Na)'s shallow NN.
5) Verification and validation (V&V) of ML algorithms using Monte Carlo simulation and metrics.

Blind comparisons are performed to evaluate the accuracy of ML-models. All models accurately estimate the Nusselt number. The maximum deviation between the ML-model predictions and CFD data is observed to be 10%. The models developed in this work provide a power tool for researchers and engineers involved in the design and the optimization of Na-cooled heat exchangers.

## 2. Methodology

1) **CFD simulations:**

Computational Fluid Dynamics (CFD) simulations are conducted to produce a dataset with 87 data points. These simulations use numerical algorithms detailed in [12] and are executed using Ansys Fluent. The objective is to obtain precise Nusselt numbers for both laminar and turbulent flow scenarios of liquid sodium in stainless steel (SS-316) rectangular miniature heat sinks with various physical properties.

The simulations encompass a range of input parameters, including heat sink width, aspect ratio, hydraulic diameter, and the Peclet number of the liquid sodium flow. The dataset generated is foundational for the development of ML models. The simulations are governed by fundamental equations for incompressible and steady-state fluid flow, including the continuity, Navier-Stokes, and energy equations for incompressible laminar flow as represented in Equations (1) to (3).

Additionally, the heat conduction equation (4), which accounts for temperature-dependent thermal conductivity, is used for the solid substrate of the heat sink.

$$\nabla \cdot (\rho u) = 0 \tag{1}$$

Equation (1) is the continuity equation for incompressible flow. It states that the mass in any given volume remains constant over time.
$\nabla \cdot$ : Divergence of a vector field.
$\rho$ : Density of the fluid (kg/m³).
u : Velocity vector of the fluid.

$$\nabla \cdot (\rho u u) = -\nabla P + \nabla \cdot \mu (\nabla u + \nabla^T u) + \rho g \tag{2}$$

Equation (2) is one of the Navier-Stokes equations, describing the velocity field of a fluid.
$\nabla P$: Gradient of pressure.
$\mu$ : Dynamic viscosity of the fluid.
$\nabla^T u$ : Transpose of the gradient of the velocity vector.
$g$ : Acceleration due to gravity.

$$\nabla \cdot (\rho c_p u T) = \nabla \cdot (k_f \nabla T) \tag{3}$$

Equation (3) is a form of the energy equation for heat transfer in fluid flow.
$c_p$: Specific heat capacity at constant pressure.
$T$ : Temperature of the fluid.
$k_f$: Thermal conductivity of the fluid.

$$\nabla \cdot (k_s \nabla T) = 0 \tag{4}$$

Equation (4) is the heat conduction equation for the solid part of the system.
$k_s$ : Thermal conductivity of the solid material.

The study employs steady-state numerical simulations with a no-slip boundary condition at the interfaces where the solid and fluid interact within the miniature heat sinks. All simulations introduce the coolant into the heat sinks at a uniform velocity and consistent temperature. Figure 1 showcases the normalized temperature and velocity distributions within a liquid sodium-cooled miniature heat sink featuring an aspect ratio of 1 and operating at a Reynolds number of 1900.

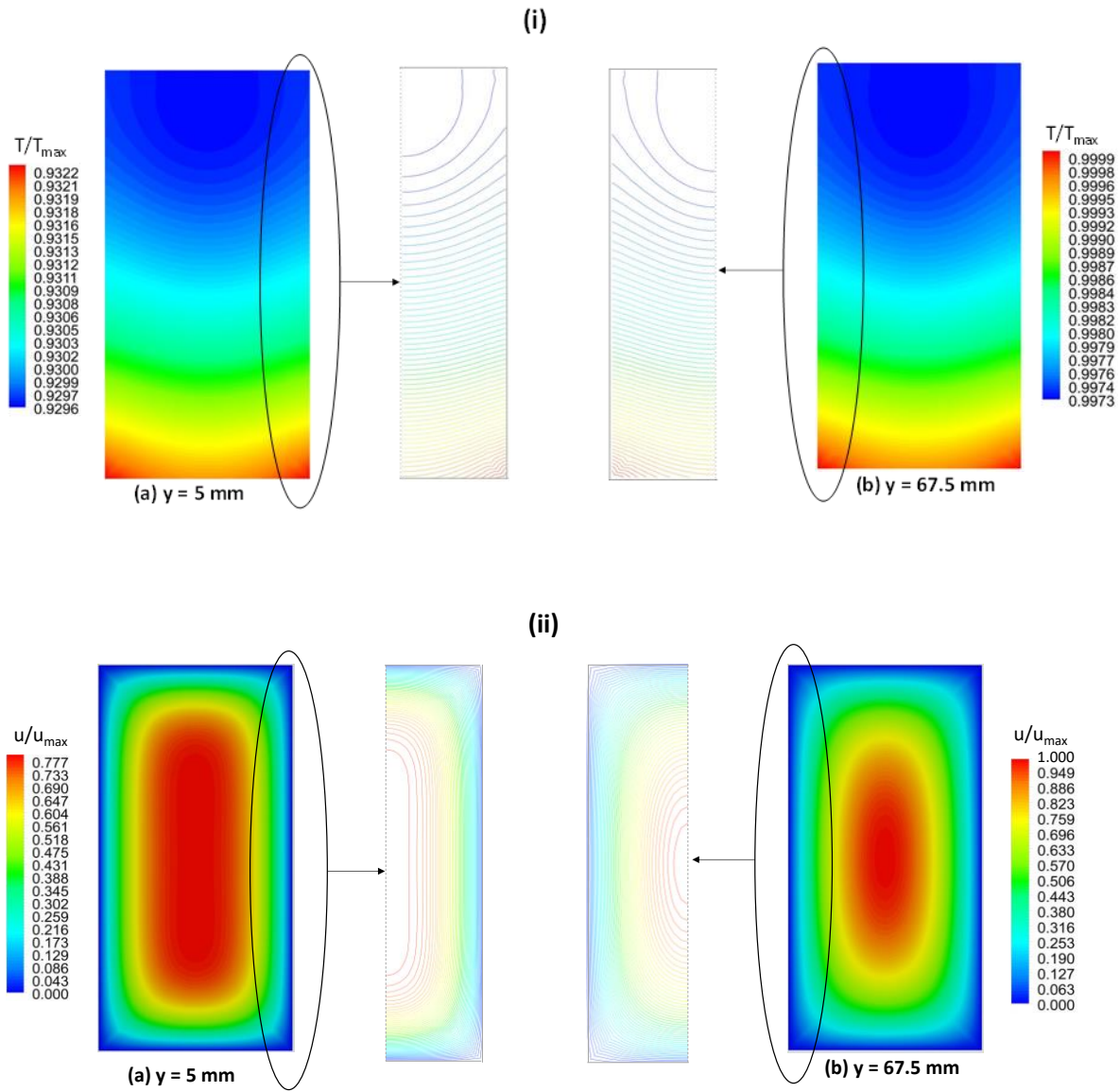

Figure 1. Normalized temperature (i) and velocity (ii) distributions at two axial locations along the rectangular heat sink for Na flow.

2) **Data preprocessing:**

The data generated from the CFD simulations are normalized. The data are divided into training, validation, and test sets to evaluate the performance of the ML algorithm.

3) **ML algorithm development:**

The ML algorithms are NN-based algorithms, GP, and SVR to estimate the Nusselt number. The algorithms are trained using the CFD training set and the performance is evaluated using the holdout dataset.

4) **Hyperparameter optimization:**

The hyperparameter of the ML algorithms are optimized using randomized search [18,19], grid search [18,19], genetic algorithm, and Besian optimization. The optimization algorithm is chosen based on the ML algorithm and its hyperparameter dimension.

5) **Error representation:**

Both general error and MAPE characterize our ML models. General error shows the absolute difference between the model predictions and the actual values, while MAPE provides the errors in percentage terms. MAPE is crucial for understanding the accuracy of the models in different situations, which is the key for designing efficient and reliable heat exchangers. N is the number of data points. The formula for MAPE is given below:

$$MAPE = (1/N) \sum |(Actual\ values - Predicted\ values) / Actual\ values| \qquad (5)$$

6) **Verification and Validation in ML:**

Verification ensures the model with hyperparameter optimization can accurately represent the training data and underlying theory. Validation confirms the model's suitability for its intended purpose, particularly using a holdout dataset.

## 2.1 Machine Learning algorithms

Three machine learning algorithms, including support vector regression (SVR), Gaussian process (GP), neural networks (NN), NN with transfer learning, and self-supervised PINN are developed and trained on a dataset of 87 data points obtained from computational fluid dynamics (CFD) simulations. The inputs of the ML algorithms are in the real vector space ($R^6$), consisting of Peclet number (Pe), the heat sink's channel hydraulic diameter ($D_h$), the channel aspect ratio ($\alpha$ = H/W), heat sink length (L), and width (W). The output of ML in the real vector space (R) is the liquid sodium average Nusselt number ($Nu_{ave}$). Due to the small size of the dataset, a 10-fold cross validation procedure is used to measure the MAPE error. As a result, 10% of dataset is used for testing and validating for each fold, and at the end, the MAPPE errors of all folds are averaged and reported as the error of the experimented ML algorithm. For the NN-based algorithms, Early stopping with 10 epochs is used.

**Optimization algorithm based on the ML algorithm**: GA is used for NN and NN with transfer learning as they have small to midsize hyperparameter. Grid-search (GS) is used for GP as it has few hyperparameter. Bayesian optimization and randomized search are used for SVR as it has midsize hyperparameter. Bayesian optimization is used for self-supervised PINN as it has midsize hyperparameter and Bayesian optimization is more robust toward the randomness of NN weights and biases initialization.

K-Fold cross-validation is used for the validation of hyperparameter structure optimization in ML algorithms with the MAPE score loss function. In K-Fold cross-validation, the dataset is divided into 'K' subsets (or folds). For each fold, the model is trained on 'K-1' subsets and validated on the remaining subset. This process is repeated 'K' times, each time with a different subset used for validation. The model's accuracy is measured with the average MAPE score across multiple folds. After determining the optimal hyperparameter, to determine the final model, a common approach is to retrain the model on the entire dataset using the optimal hyperparameter without using K-Fold cross-validation. Due to randomness in NN, the NN Hyperparameter is validated using Monte Carlo simulation as show in Figure 2. After obtaining the optimal hyperparameter, Monte-Carlo simulation with 500 evaluations in the validation phase trains the NN at each iteration and measures the metrics of variation in prediction for the holdout dataset, MAPE, and epochs. These metrics validate the optimal hyperparameter for the implementation to obtain the predictions.

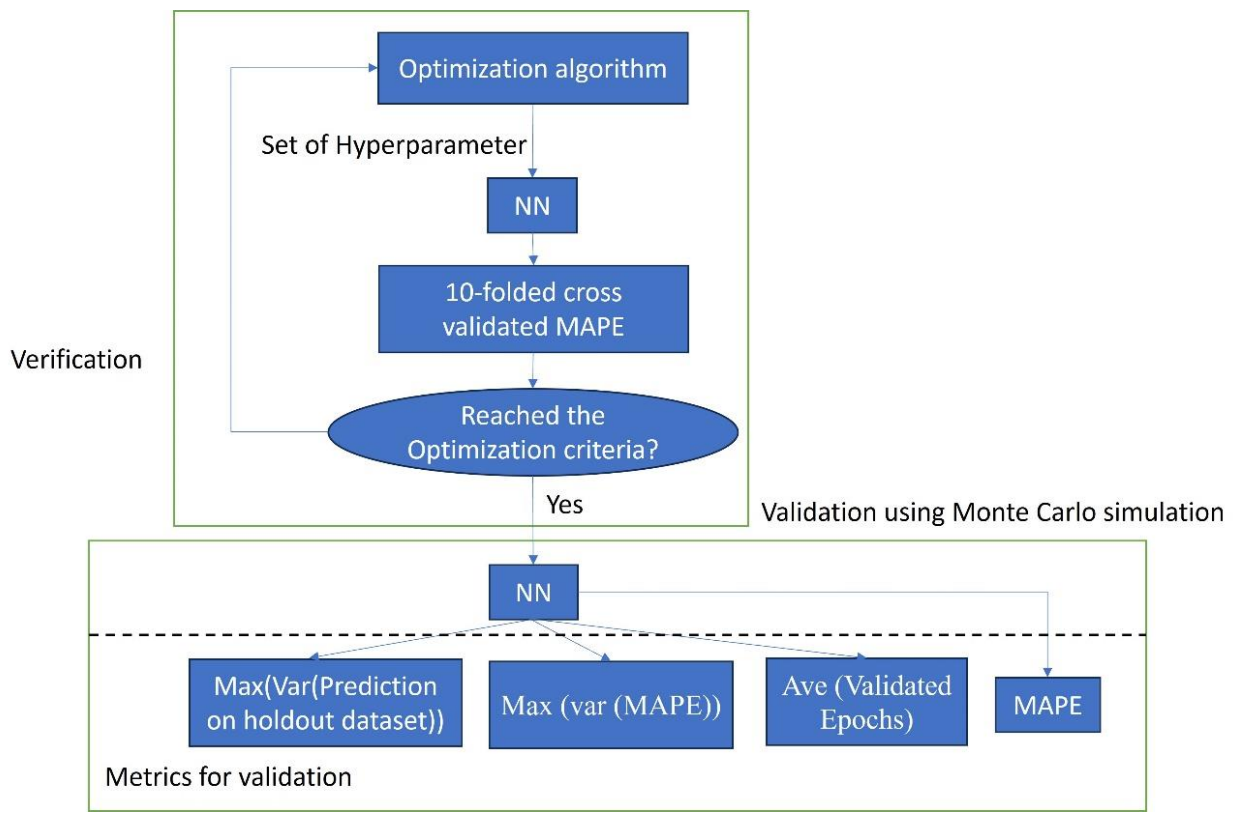

Figure 2. V&V of optimal hyperparameter in NN-based ML algorithms.

### a) Support vector regression (SVR)

The SVR hyperparameter that are tuned for minimizing the MAPE criterion include the kernel, gamma, epsilon, and C. The hyperparameter are characterized as follows:

**C**: The penalty parameter of the error term in SVR, varied between 1e-6 and 1e+6 on a log-uniform scale.

**Gamma(γ)**: A parameter for non-linear kernels, varied on a log-uniform scale between 1e-4 and 1e+1.

**Epsilon(ε)**: This parameter specifies the epsilon-tube within which no penalty is associated in the training loss function with points predicted within a distance epsilon from the actual value, also varied on a log-uniform scale.

**Kernel function**: Specifies the kernel type to be used in the SVR. The options are 'rbf', 'linear', and 'sigmoid'. These are categorical variables, meaning the optimizer will choose among these discrete options.

Optimal hyperparameter are obtained using a random search optimization given in Appendix A. This approach involves fitting the SVR model for each possible combination of hyperparameter and selecting the combination that achieves the best model performance based on 10-fold cross-validated MAPE criterion. The SVR architecture and algorithm parameters are given within a list 'P' for the search-space. The list's index 'i' operates as the optimization variable. RS identifies the optimal configuration $P_{opt}$, solving the hyperparameter optimization problem. RS uses a uniform distribution to randomly select 'i'. The process terminates when the iteration count 'k' reaches the predefined value $T_{Iter}$. During each RS iteration, the SVR is trained and the corresponding MAPE is calculated. When the decrease in the current MAPE from the previous MAPE exceeds a predetermined threshold $T_{Drop}$, $T_{Iter}$ is incremented by a constant $k_{Iter}$ and the variable Span. Initial values for $T_{Iter}$, MAPE, Span, $k_{Iter}$, and $T_{Drop}$ are set to ensure convergence with minimal iterations. This strategy increases the maximum number of search iterations provided MAPE reduction surpasses $T_{Drop}$. If MAPE converges to the optimal value and no further reductions greater than $T_{Drop}$ are observed, the termination criterion is fulfilled. The RS process is outlined in Algorithm 1. This method balances computational efficiency and optimization. Bayesian optimization is also applied to SVR to find the optimal hyperparameter with 10-fold cross-validated MAPE. The hyperparameter search space is as follows.

### b) Gaussian Process (GP)

The kernel function is selected from a set including the Linear Dot Product (LDP), Exponential Sine Square (ESS), Matérn, Radial Basis Function (RBF), and Rational Quadratic (RQ). The covariance function in GP is a linear mixture of the kernel function, a constant term, and a white noise element. The kernel function is chosen to obtain the smallest MAPE. Grid search is used to define the optimal kernel using the MAPE criterion due to the small number of hyperparameter combinations.

### c) Neural Network (NN)

NNs consist of layers of interconnected "neurons," which are mathematical functions that process information. The layers of NN are organized into input, output, and hidden layers. The number of layers varies between 1 to 3, and the number of neurons at each layer varies between 1 to 10. The activation function for the hidden layer is logistic sigmoid transfer function, and the activation function for the output neuron is Linear transfer function (Purelin). logistic sigmoid activation function is given as:

$$f(x) = \frac{1}{1 + exp(-x)} \tag{6}$$

Purelin function is as follow:

$$f(x) = x \tag{7}$$

Where x is the input to a neuron. The structural hyperparameter of NN is the number of layers and number of neurons. The initial values for weights and bias of the neurons are chosen randomly with a uniform distribution between (0,1). The parameters weight and bias of the NN are optimized through the Levenberg–Marquardt algorithm (LMA) and stochastic gradient descent (SGD). The optimization criterion is MAPE obtained by LMA and SGD. SGD updates the model's parameters in small steps in the direction that reduces the model's error most rapidly, defined by the gradient of the loss function. SGD estimates the gradient based on a single randomly selected or sequentially accessed example (or a small batch of examples) at each step. LMA is designed to approach the problem with gradient descent when the parameters are far from their optimal value and then shift towards the Gauss-Newton method as the parameters get closer to the optimum. This combination gives LMA robustness without sacrificing speed or accuracy.

LMA provides lower MAPE compared to SGD for the parameters' optimization, as LMA is efficient and designed for shallow NN. GA optimization criteria for optimization is 10-fold cross-validated MAPE, and the GA optimization parameter is to reach 500 generation or drop of 10-fold cross-validated MAPE for 1e-6.

d) **Transfer Learning**

Transfer learning is a machine learning technique where a model developed for one task is initializing a model on a second task. Transfer learning is leveraged to enhance the performance of ML models in CFD simulations, which is measured in accuracy, computation consumption, and robustness. Our aim is to determine the optimal number of layers transfer from water dataset, a frequently used coolant, to liquid sodium (Na) dataset, a coolant with limited experimental data, using GA. The parameters weight and bias of the NN in the transfer learning are optimized through the LMA.

The current data and its corresponding NN structure for water is based on a previous paper [14] with 3 neurons in the first hidden layer, 2 neurons in the second hidden layer, and 1 output neuron. MAPE for water is 0.052. 10-fold cross-validation is used to minimize bias in MAPE for training. GA discovers the optimal structure of the NN for Liquid Sodium (Na) and the optimal transfer learning approach for transferring layers from water to Liquid Sodium (Na). For the Liquid Sodium (Na) dataset, the NN is initialized as follows: the first layers are initialized using weights from the first layers of water dataset, where the number of layers transferred from water's NN is determined by GA. The remaining hidden layers are initialized randomly with a uniform distribution between (0,1) for the neurons. The performance is evaluated with 10-fold cross validated MAPE as the performance metric.

e) **Self-Supervised Physics Informed Deep Learning**

A key component of PINN is the incorporation of a physics-based model alongside the neural network. This model encapsulates the relationship between input features (such as Alpha, L/D ratio, Peclet number) and the Nusselt number. By incorporating this physics-based model, the network is informed by both the empirical data and the underlying physics, leading to a more robust and accurate predictive model.

The PINN model uses PyTorch framework and CUDA GPUs A100 for parallel training. A distinctive aspect of the proposed PINN architecture is the addition of a dedicated neroun for computing a physics coefficient, which integrates physical laws into the learning process. This integration ensures that the model's predictions are not only data-driven but also adhere to physical laws. The physics prediction of Nusselt number $\widehat{Nu}_{ave}$ [12, 17] is given as the following.

$$\gamma = \left[ \frac{\alpha^{0.35}}{\left(\frac{L}{D_h}\right)^{0.1}} \right] \quad (8)$$

$$Pe^* = Pe\,\gamma \quad (9)$$

$$Nu^* = 0.164 + 10.2\gamma - 12\gamma^2 \quad (10)$$

$$\widehat{Nu}_{ave} = Nu^*(1.0 + 0.135(Pe^*)^{0.388}) \quad (11)$$

The loss function of self-supervised PINN is given as

$$Loss = MAPE(prediction, labels) + MSE(physics\ prediction, physics\ coefficient) \quad (12)$$

Where mean square error (MSE) is given as:

$$MSE = (1/N) \sum |(Actual\ values - Predicted\ values)^2| \quad (13)$$

N is the number of data points. MSE is chosen for the physics loss function as physics coefficient better maps between 0 and 1 during the training. The architecture comprises 3 fully connected layers, with the number of neurons in each layer optimized through Bayesian optimization. The number of neurons at each layer varies between 2 to 20. The stopping criterion for optimization is the change in the 10-fold cross-validated MAPE with a 0.001 threshold in 50 iterations or 100 iterations in total. The activation function for the physics coefficient neuron layer is logistic sigmoid transfer function, and the neuron's activation function of the $Nu_{ave}$ prediction and hidden layers are Rectified Linear Unit (RELU). This dual approach leverages the strengths of both activation functions in their respective roles. The logistic sigmoid activation function is given as:

$$f(x) = \frac{1}{1 + exp(-x)} \quad (14)$$

RELU function is as follow:

$$f(x) = \max(x, 0) \quad (15)$$

Where $x$ is the input to a neuron. The weight and bias of the NN are optimized through Adaptive Moment Estimation (Adam), Root Mean Square Propagation (RMSprop), and SGD. The optimization criterion is MAPE. Figure 3 shows the structure of the self-supervised PINN and hyperparameter optimization.

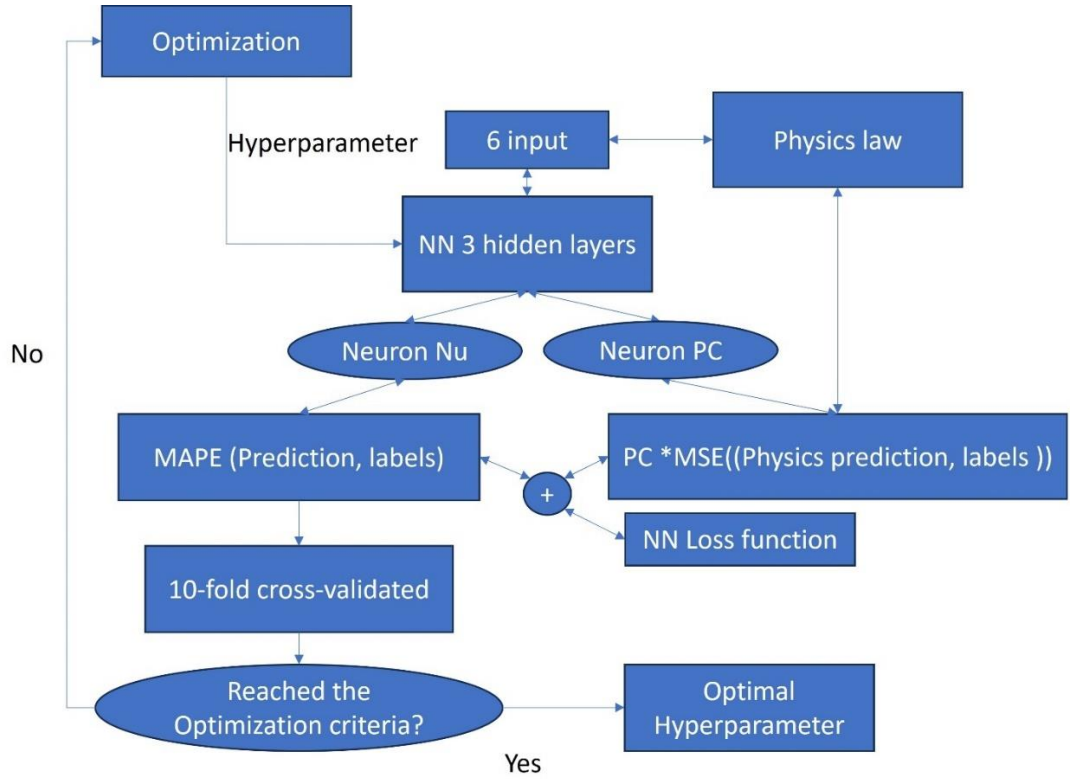

Figure 3. Self-supervised PINN, PC represents the physics coefficient, mean square error (MSE) is used for the loss function of the physics estimation. Two-sided arrows show the backpropagation.

The summarized optimization algorithms as a pseudocode is given in Appendix B.

1. **Data Analysis**

The Kernel Density Estimate plot distribution of the 87 dataset is shown in figure 4. The Pe number approximately shows a Gaussian distribution.

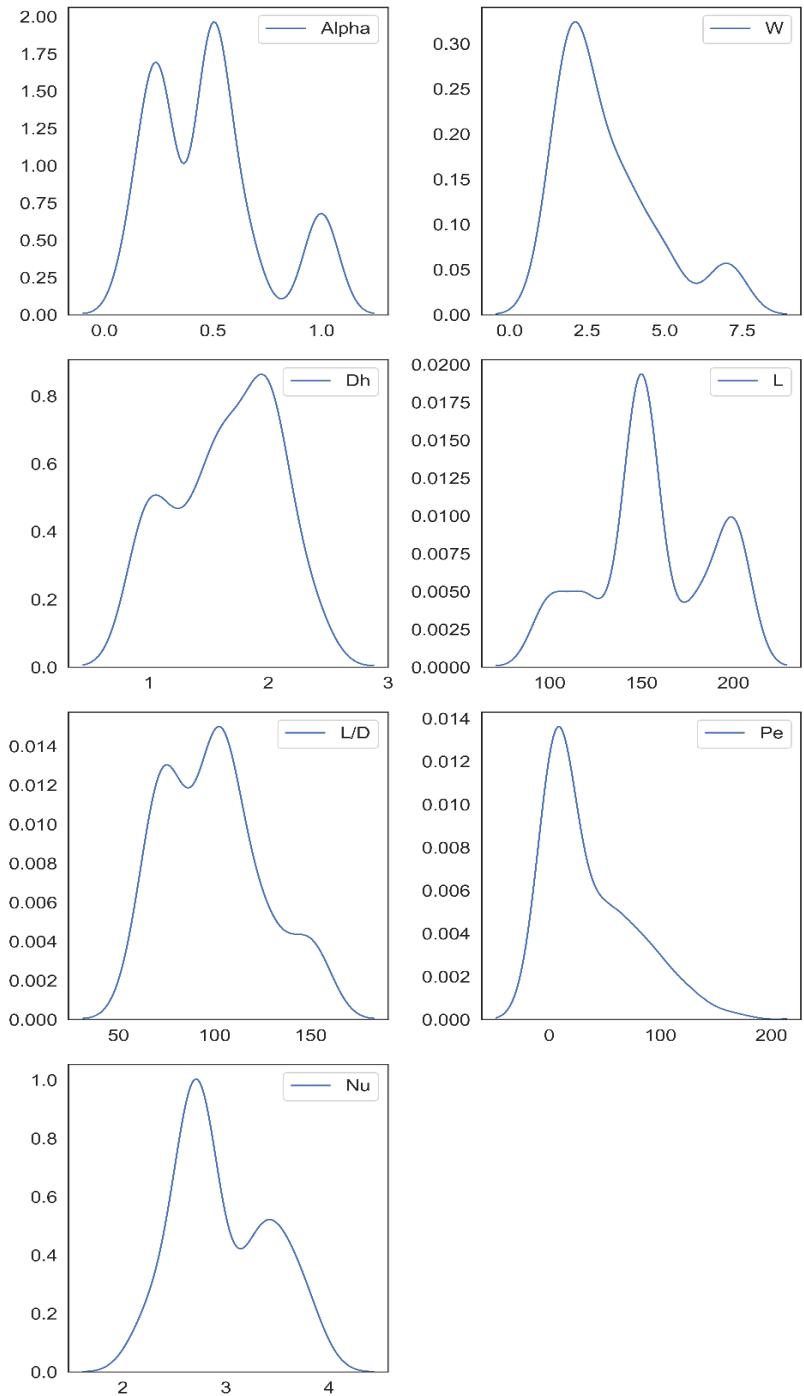

Figure 4. The Distribution of α, W, Dh, L, L/D, Pe, $Nu_{ave}$ for the data are obtained by CFD for the flow of Liquid Sodium (Na) within miniature heat sinks. Each subplot in the figure depicts the Kernel Density Estimate plot for each variable. The KDE plot is a smoothed histogram.

The data is positive and finite. The statistics for the data are presented in Table 1. $P_k$ represents the kth percentile of a variable x such that k% of the values in the distribution of that variable are less than or equal to the corresponding value of Tr in $P_k(x)= Tr$

Table 1. Dataset statistics.

| y | Mean | Mode | Variance | Max | $P_{99}$ | $P_1$ |
|---|---|---|---|---|---|---|
| α (H/W) | 0.4752 | 0.5000 | 0.0671 | 1 | 1 | 0.1428 |
| W (mm) | 3.1551 | 2 | 2.5599 | 7 | 7 | 1.5000 |
| Dh (mm) | 1.6464 | 2 | 0.1784 | 2.3300 | 2.3300 | 1 |
| L (mm) | 155.9770 | 150 | 984.7901 | 200 | 200 | 100 |
| L/D | 99.2258 | 75 | 633.6310 | 150 | 150 | 75 |
| Pe | 36.0843 | 3.9066 | 1526.7608 | 162.7780 | 130.2220 | 3.9066 |
| $Nu_{ave}$ | 2.9661 | 2.1986 | 0.2002 | 3.8500 | 3.8262 | 2.5100 |

Table 1 shows that the mean of the data varies from approximately 0.5 to 160. The largest difference in max and $P_{99}$ is for the Pe number, which is visible in figure 3. In NN, this data is normalized for training and testing.

## 4. Results and Discussion

In this section, the results of the given methods are shown and the bench marking section compares the results.

### 4.1 Kernel-Based Machine Learning

As shown in table 2, Gaussian Process (GP) and Support Vector Regression (SVR) methods have the optimal value when using Radial Basis Function (RBF) kernel. SVR demonstrates higher accuracy due to data distribution and having more hyperparameter for extrapolation. Figure 5 illustrates the variation of MAPE against the number of iterations when MAPE decreases by more than $T_{Drop}$. The randomized search (RS) method reaches a minimum MAPE of 2.72, plateauing at iteration 4. Bayesian optimization, which had a stopping criterion set at 8 iterations, converges faster in 17 iterations, nearly twice as quickly as RS.

Table 2. Comparison of performance between GP and SVR

| Method | Kernel | MAPE |
|---|---|---|
| GP | RBF | 0.0756 |
| SVR-RS | RBF | 0.0272 |
| SVR-Bayesian | RBF | 0.0125 |

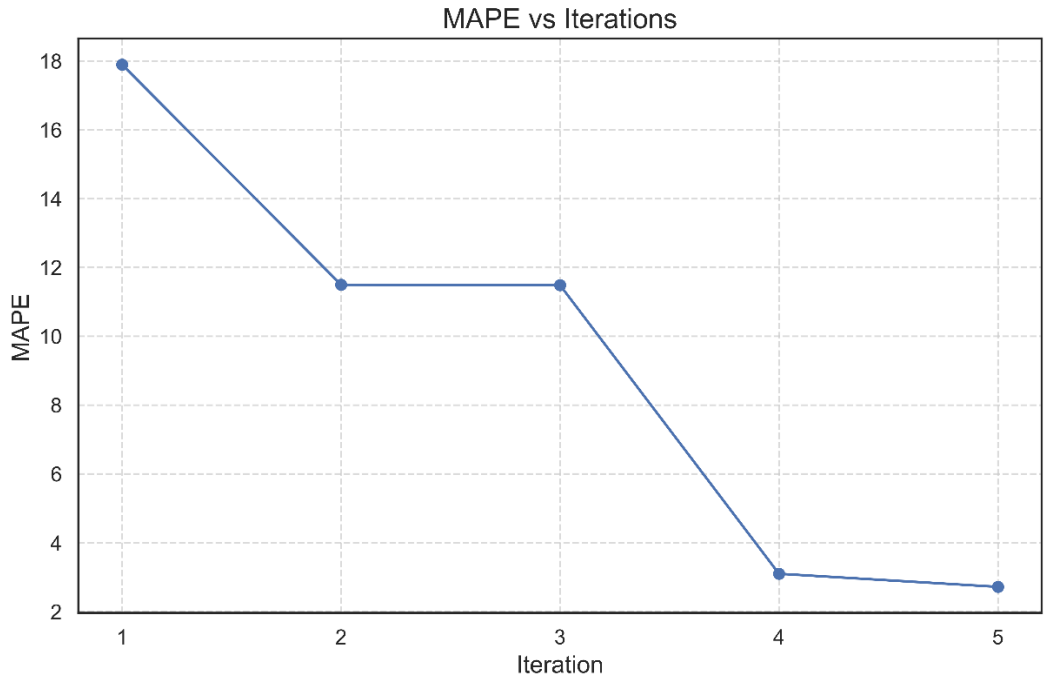

Figure 5. MAPE variation as a function of iteration in RS optimization when MAPE drops by more than $T_{Drop}$.

Table. 3 shows the hyperparameters of SVR. The optimal value of the regularization parameter C in SVR is 10 for RS and 27.23 for Bayesian optimization.

Table 3. Optimal MAPE using RS and Bayesian optimization for SVR.

| Optimization algorithm for SVR | Kernel | C | $\gamma$ | $\varepsilon$ | MAPE |
|---|---|---|---|---|---|
| RS | RBF | 10 | 0.0001 | 0.001 | 0.0272 |
| Bayesian | RBF | 27.23 | 0.0007 | 0.0031 | 0.0125 |

## 4.2 Transfer Learning and neural networks

Table 4 and Figure 6 show that the optimal transfer method obtained by GA achieves a MAPE of 0.0020, whereas the error without transfer is 0.0028. This optimal MAPE for transfer learning is obtained when the first layer of water NN is transferred. The optimal number of hidden layers and neurons for transfer learning is 2 hidden layers, with 3 neurons from the first layer of water and 8 neurons initialized between 0 and 1 randomly for the second hidden layer as determined by GA. Without transfer learning, the optimal structure is 1 hidden layer with 8 neurons. Figure 3 illustrates that transferring the first layer results in the lowest MAPE error, capturing general information, while transferring layers closer to the output results in higher errors, indicating more specificity to the water fluid domain of data.

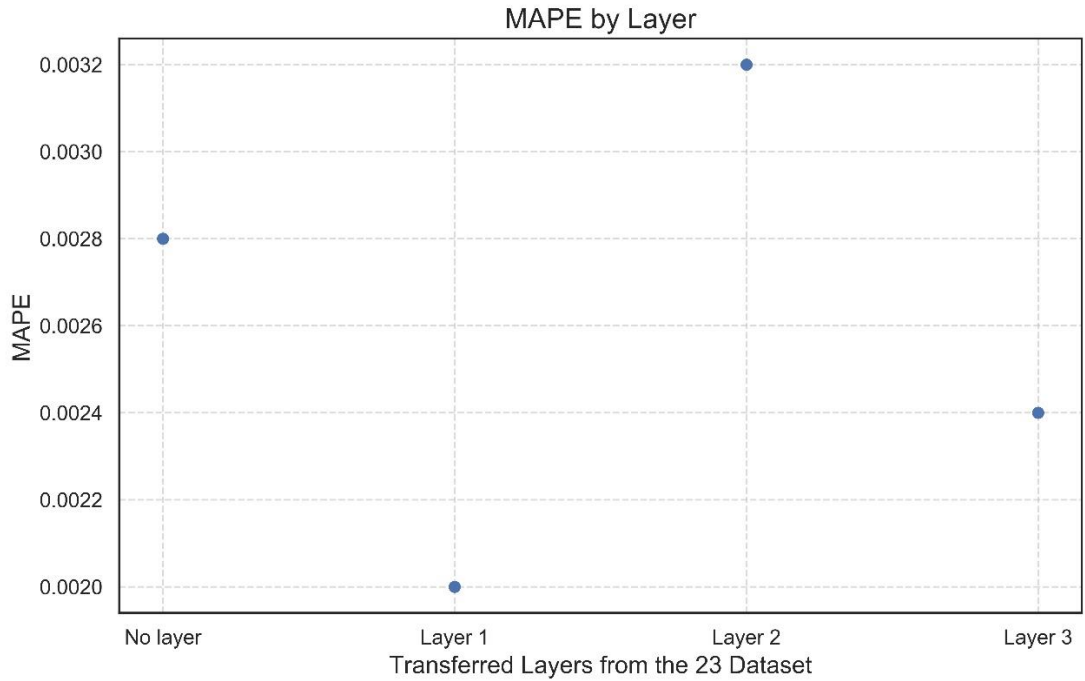

Figure 6. MAPE error for different transfer layers. The first layer transfer has the lowest error, and the transfer before the output has the highest error.

Figure 8 compares the Nusselt number ($Nu_{ave}$) between water and liquid sodium (Na) using boxplots. The boxplots illustrate the distributions of $Nu_{ave}$ for both fluids, showing a wider spread for liquid sodium (Na) compared to water. The medians of the Nusselt numbers for water and liquid sodium (Na) do not overlap at the notches, indicating a statistically significant difference between the two fluids [21]. This difference suggests the rejection of the null hypothesis. The Mann-Whitney U test [22], also known as Wilcoxon rank sum test, can be used to drive the P-value. Unlike the t-test, the Mann-Whitney U test does not assume a normally distributed dataset. Figure 7 shows the distribution of water (shown in blue) and Na (shown in orange) using Kernel Density Estimates (KDE). The x-axis represents the data values, and the y-axis represents the density of these values. The P-value is much less than 0.05, leading to the rejection of the null hypothesis which indicates a significant difference between the fluids. The $Nu_{ave}$ is considered a metric and indication for transferring the NN from water to liquid sodium (Na), as the knowledge of the $Nu_{ave}$ from water is applied to liquid sodium (Na). As a result, the first layer in water NN, which generalizes information the most, is best suited for transfer based on the metric of Na distribution.

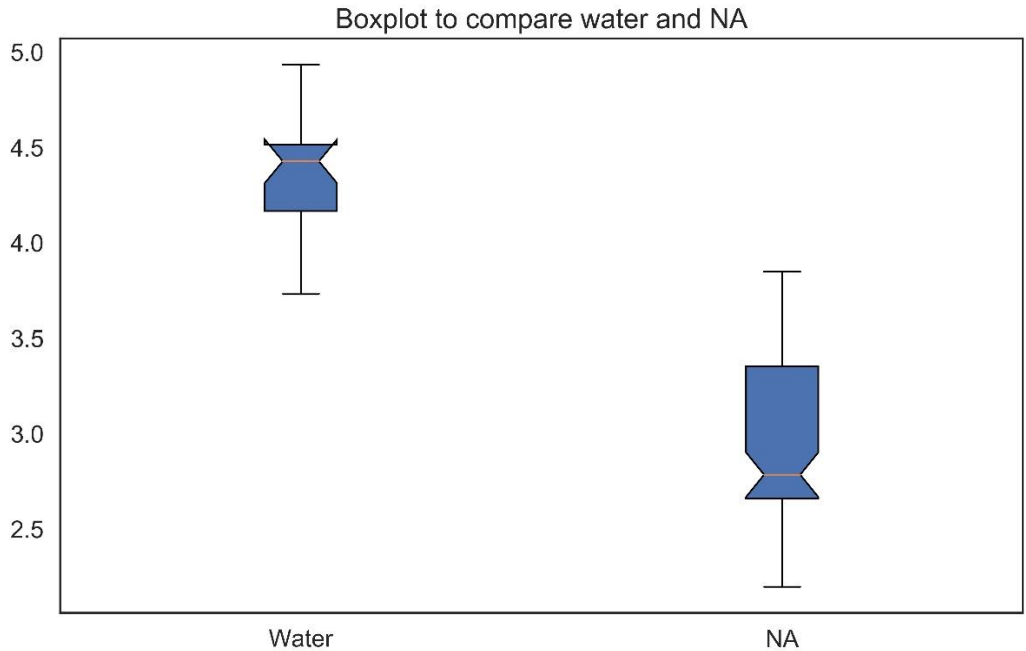

Figure 7. Notched boxplot comparing Nusselt numbers of water and liquid sodium (NA). The median $Nu_{ave}$ of water is higher than that of NA, and the Water dataset has less variability

Table 4. Performance metrics for transfer learning

| Transfer Method | Error (MAPE) |
| --- | --- |
| Optimal Transfer | 0.0020 |
| No Transfer | 0.0028 |

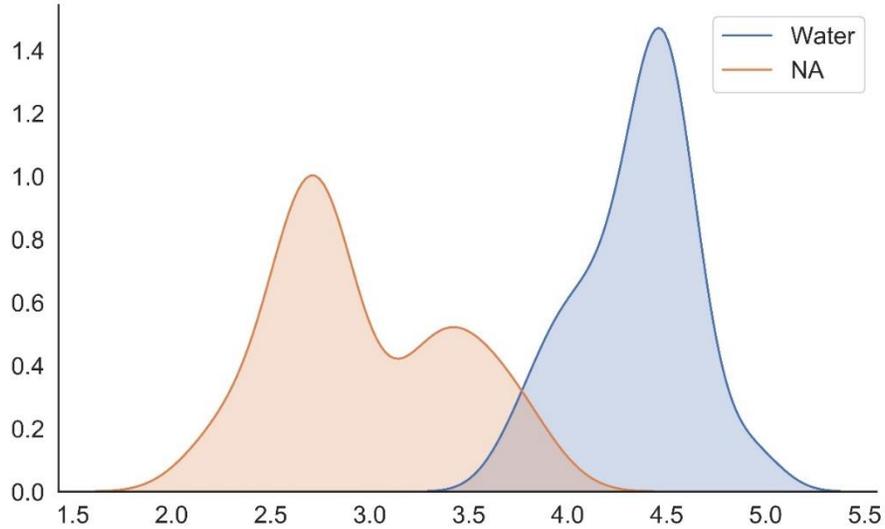

Figure 8. Kernel density estimates (KDEs) of water and liquid sodium (NA) for Nusselt number. The water KDE is taller and narrower than the NA KDE, indicating that the water data has less variance.

### 4.3 Self-supervised physics informed deep learning.

In Self-supervised PINN, Bayesian optimization finds the optimal number of neurons as 20 for the first 2 layers and 12 for the last hidden layer. Adam optimizer is obtained with a learning rate of 0.34. The 10-fold cross-validated MAPE is 0.0185, obtained during 100 iterations of Bayesian optimization. Figure 9 Shows the distribution of physics coefficient neuron, centered around 0.5.

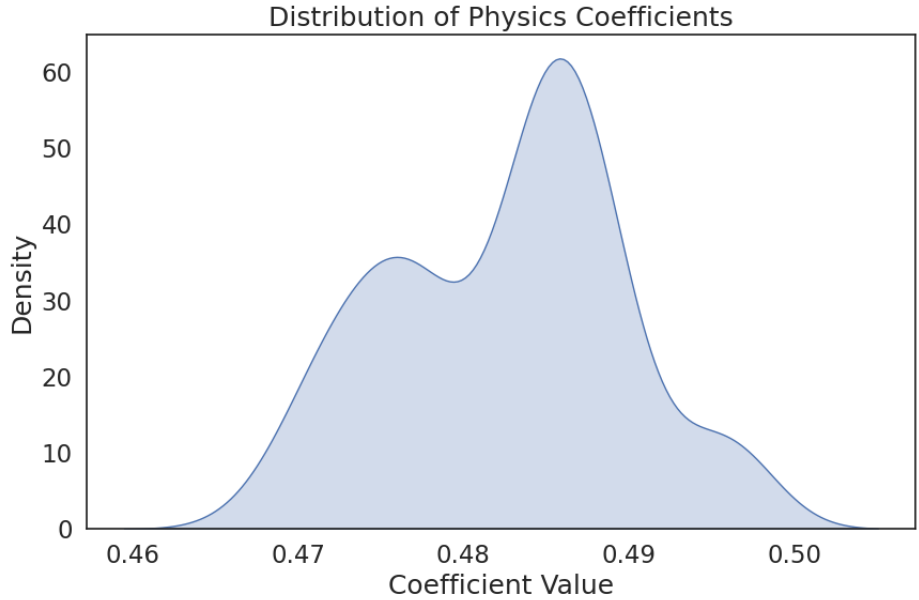

Figure 9. Physics coefficient distribution

### 4.3 Benchmarking and Validation

Table. 5 compares MAPE for different ML methods, where SVR-Bayesian provides the lowest MAPE and GP relatively the highest MAPE for Nusselt number ($Nu_{ave}$) estimation without considering the holdout dataset.

Table 5. MAPE values for NN, GP, and SVR.

| Error | NN with Transfer Learning | NN | PINN | GP | SVR-RS | SVR-Bayesian |
|---|---|---|---|---|---|---|
| MAPE | 0.002 | 0.0028 | 0.0185 | 0.0756 | 0.0272 | 0.0125 |

Using 500 Monte Carlo simulation, for a set of obtained hyperparameter by the optimization, the predictions and MAPE for the NN is shown in Table 6.

Table 6. Metrics for NN, GP, and SVR. "var" represents variance, and "Ave" shows the average. "TR_NN" shows the transfer learning approach.

| Validation metrics | TR_NN | NN | PINN |
|---|---|---|---|
| Max (var (predictions (holdout dataset))) | 0.2183 | 0.2323 | 0.11461 |
| Max (var (MAPE)) | 0.0002 | 0.0002 | 0.0008 |
| Ave (Validated Epochs) | 14.2972 | 14.1990 | 35 |

Table 6 shows that PINN has half the variation in the prediction of holdout dataset, while it has four times higher Max (var (MAPE)). This indicates that although the variation of MAPE is high in the training phase, the estimation is robust toward this uncertainty in the holdout dataset. It validates the self-supervised PINN toward hyperparameter optimization and robustness toward predictions. In the self-supervised PINN, the number of epochs is approximately 3 times higher for the training, partially due to using LMA backpropagation in NN and transfer learning. This epoch number is measured during the early stopping phase of cross-validation since it defines the computation requirement for the number of epochs in the hyperparameter optimization.

To validate the ML algorithms, a holdout dataset that is not seen by the ML models is used to test their performance. Figures 10 show the estimation for the holdout dataset using GP and SVR. The estimation is mostly within an 8% margin of error. GP shows the highest error indicating potential underfitting due to the low number of hyperparameter to tune based on the complexity of data shown in figure 4. Figure 11 shows that the NN-based algorithms also provide estimations within mostly an 8% margin of error, with self-supervised PINN showing the lowest margin of error. Overall, Figures 10 and 11 show that all ML methods provide fairly accurate estimations for Na Nusselt numbers in stainless steel miniature heat sinks. Figure 11 shows that only self-supervised PINN remains within the 8% range. Self-supervised PINN demonstrates higher robustness toward the weight randomness of NN with the validated optimal hyperparameter.

In [12], it was shown that using the physics given in Eq. (11) reaches a margin of 5% to 10% margin of error. This indicates that the physics multivariate regression in Eq. (11) could marginally reduce the error of ML. However, the physics in Eq. (11) still has a margin of 5% to 10% error range, so the best approach is to combine the data and physics to balance the physics analysis and ML estimation of data for a more robust and reliable prediction. This approach is similar to ensemble learning, where multiple ML methods combine their estimation for robust and accurate estimations.

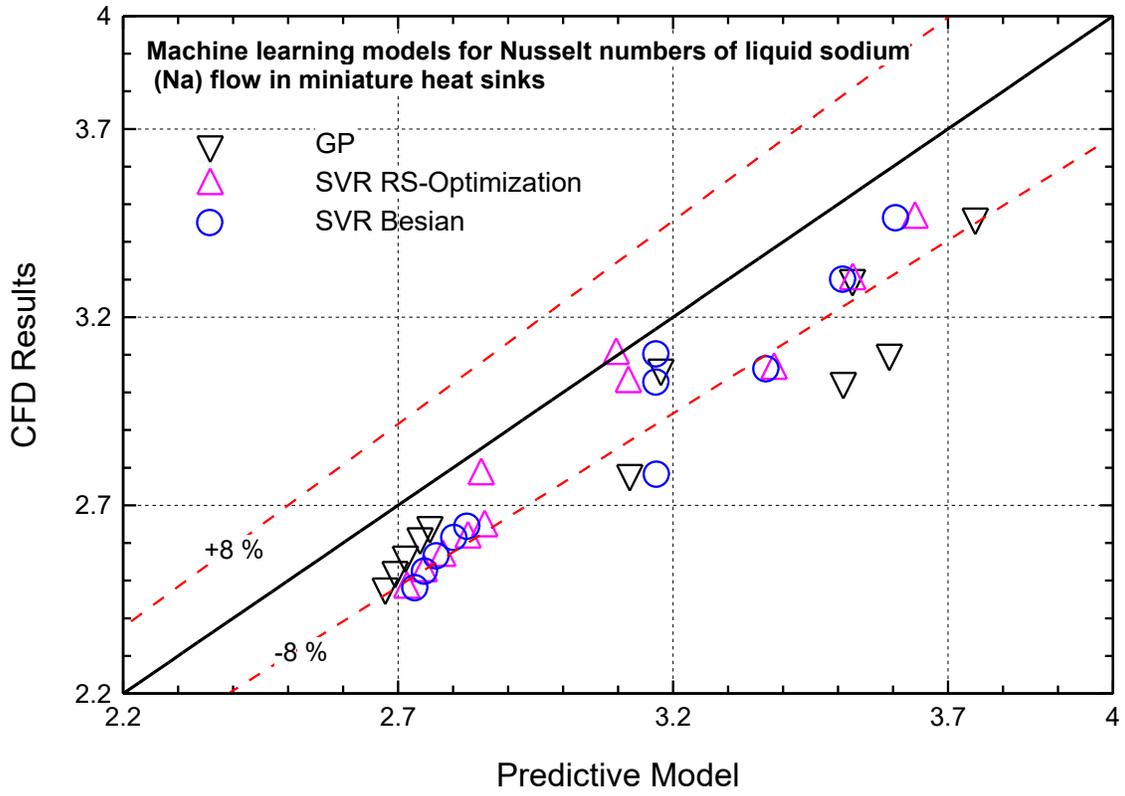

Figure 10. Kernel-based ML algorithms

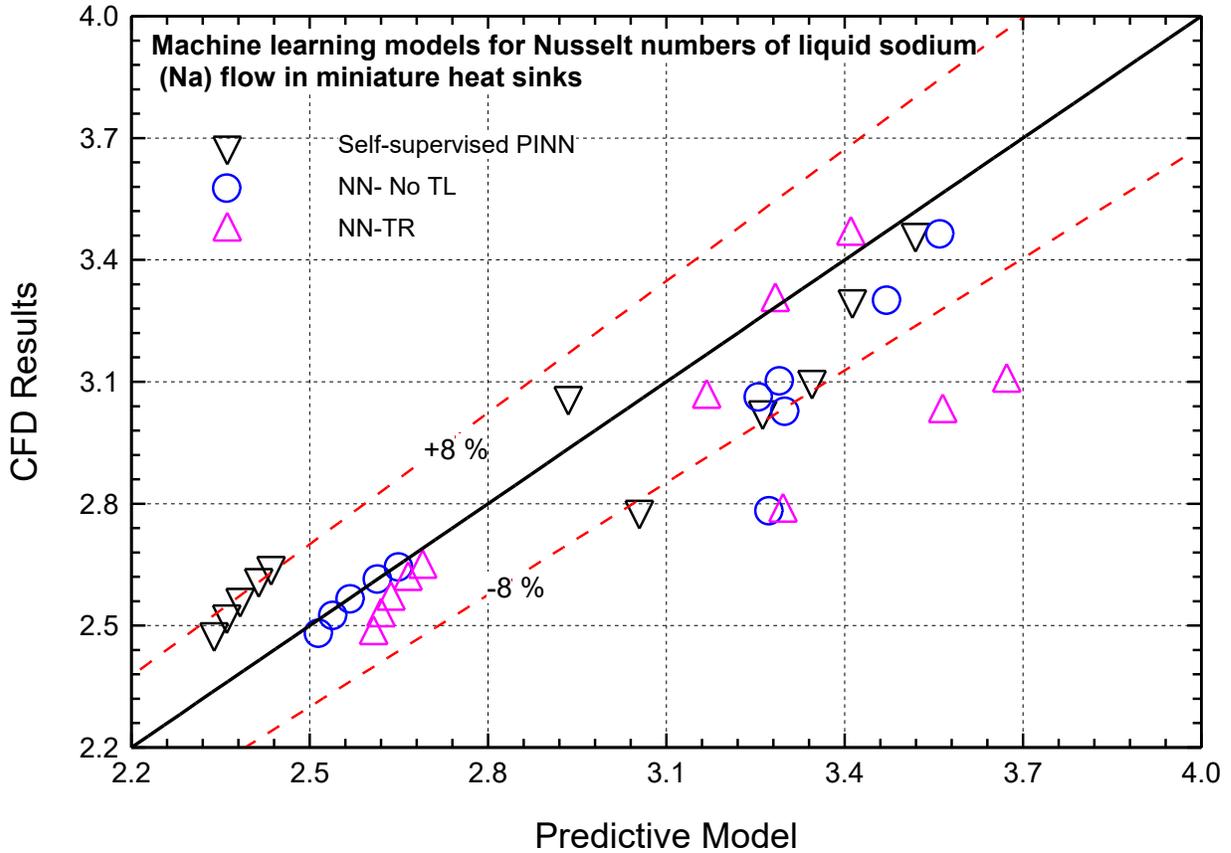

Figure 11. NN-based algorithms error using holdout dataset

## 5. Conclusion

Experimental investigations and high-fidelity CFD simulations for Nusselt number calculation are often expensive and time-consuming, particularly for applications involving liquid metals under extreme working conditions where experimental data are often rare or unavailable. As an alternative, this research investigates the effectiveness of using Gaussian Process (GP), Support Vector Regression (SVR), and NN-based learning algorithms to predict the Nusselt number for liquid sodium (Na) within miniature heat sinks. Despite the small amount of data (87 data points) used for training, our machine learning (ML)-based approach results are mostly within an 8% error margin for the holdout dataset, with self-supervised PINN achieving an error strictly within 8%.

Kernel-based machine learning models, such as SVR and GP, successfully estimate the Nusselt number with relatively high accuracy. Notably, the transfer learning approach demonstrated that neural networks trained on data from a different fluid domain (water) could be effectively adapted to the Na domain, marginally enhancing estimation accuracy (MAPE) for the Nusselt number. Considering the distribution of Nusselt number as a metric and indication of similarity between water and Liquid Sodium (Na), the first layer transfer provided the most benefit for Na Nusselt number estimation, enhancing NN extrapolation.

Moreover, self-supervised PINN demonstrated greater robustness and accuracy, leveraging both empirical data and underlying physics to achieve reliable predictions. The integration of physics-informed methods within the ML framework not only improved accuracy but also ensured that the model's predictions adhered to physical laws. This hybrid approach offers a more efficient and scalable alternative to traditional physics-based regression, which often relies on rigorous trial-and-error processes. The ML models are easier to train and require significantly less computational resources, while achieving better or comparable accuracy.

Beyond the specific application to heat transfer in miniature heat sinks, the PINN method introduced in this study provides a flexible and powerful framework that can be extended to other fields of research, including broader computational fluid dynamics (CFD) applications. By incorporating domain-specific physics into ML models, this approach can enhance the predictive capabilities of simulations in complex and data-scarce environments. Researchers and engineers across various disciplines can leverage this method to enhance their design and optimization processes, leading to more innovative and efficient solutions in fields ranging from aerospace engineering to energy systems.

The developed ML algorithms, verified and validated through rigorous cross-validation and Monte Carlo simulations, provide powerful tools for researchers and engineers involved in the design and optimization of Na-cooled heat exchangers, enhancing the development process and potentially leading to more innovative solutions in heat transfer applications.

## Appendix A

Algorithm 1: Randomized search optimization.

1) Initialize the optimization parameters and iteration index k = 0;
2) Randomly choose the initial value $i_0$ and define MAPE($P_{i0}$);
3) Generate the integer $V_{k+1} \in [1,|P|]$ randomly with a uniform distribution ;
4) If MAPE($P_{Vk+1}$) < MAPE($P_{ik}$)

   $i_{k+1} = V_{k+1}$

   Otherwise $i_{k+1} = i_k$

   End;
5) If MAPE($P_{ik}$)−MAPE($P_{Vk+1}$) > $T_{Drop}$

   Increment *Span*

   $T_{Iter} = k_{Iter} \times Span + T_{Iter}$

   End;
6) If k reaches $T_{Iter}$, stop. Otherwise, increment k and return to Step 3.

## Appendix B

Algorithm 2: Summarized ML and Optimization Algorithms

1. Initialize parameters and datasets

- Define the dataset from CFD simulations

- Normalize the data

- Divide the data into training, validation, and test sets

2. Machine Learning Model Selection

   - Select NN, NN with transfer learning, SVR, GP, and Self-Supervised PINN models based on the dataset

   - Define model-specific hyperparameters and initialize them

3. Hyperparameter Optimization

   - For NN and NN with transfer learning:

     a. Use Genetic Algorithm (GA) to optimize the number of layers and neurons

     b. Apply Levenberg–Marquardt Algorithm (LMA) or Stochastic Gradient Descent (SGD) to optimize weights and biases

   - For GP:

     a. Use Grid-Search (GS) to determine the optimal kernel function

   - For SVR:

     a. Apply Bayesian optimization and Randomized Search for kernel, gamma, epsilon, and C hyperparameters

   - For Self-Supervised PINN:

     a. Implement Bayesian optimization to find the optimal number of neurons and layers

4. Cross-Validation and Model Validation

   - Perform K-Fold Cross-Validation (K = 10) for each model:

     a. Divide the dataset into 'K' subsets

     b. Train the model on 'K-1' subsets and validate on the remaining subset

     c. Repeat the process 'K' times with different subsets

d. Compute the average MAPE score across all folds

   - After finding the optimal hyperparameters, retrain the models on the entire dataset

5. Monte Carlo Simulation for Neural Networks

   - Initialize Monte Carlo Simulation with 500 iterations

   - For each iteration:

     a. Train the NN model using the optimal hyperparameters

     b. Measure and record MAPE, variation in prediction, and epochs

   - Validate the NN model by analyzing the distribution of MAPE and prediction variations

6. Final Model Deployment

   - For the final model:

     a. Retrain using the entire dataset and optimal hyperparameters without K-Fold Cross-Validation

     b. Deploy the model to predict the Nusselt number for the miniature heat sinks

7. Output Results and Metrics

   - Report the final model performance, including MAPE, epochs, and prediction variation

   - Compare models and select the most robust one based on the validation metrics

8. Conclude with Model Verification and Validation (V&V)

   - Validate the final model with a holdout dataset and assess accuracy and robustness

   - Document the process for reproducibility and future work

**Acknowledgment**

The author thanks Prof. Nima Fathi and Prof. Mahyar Pourghasemi for valuable guidance and discussions during the stages of this project. Parts of this work are part of a broader collaborative effort, and a revised version including co-authors is in preparation for journal submission.